\pdfoutput=1

\documentclass[11pt]{article}

\usepackage{acl}

\usepackage{times}
\usepackage{latexsym}
\usepackage{float}
\usepackage{amssymb}
\usepackage{amsmath}
\usepackage{graphicx}
\usepackage{caption}
\usepackage{subcaption}
\usepackage{booktabs}
\usepackage{algorithm}
\usepackage{algorithmic}
\usepackage{hyperref}
\usepackage{makecell}
\usepackage{tikz}
\usetikzlibrary{arrows,automata}

\usepackage{arydshln}

\graphicspath{ {./images/} }

\usepackage[T1]{fontenc}

\usepackage[utf8]{inputenc}

\usepackage{microtype}

\title{Policy Compliance Detection via Expression Tree Inference}
\newcommand{\affilsup}[1]{\rlap{\textsuperscript{\normalfont#1}}}

\author{
    Neema Kotonya  \affilsup{1,2} \qquad
    Andreas Vlachos \affilsup{1,3} \qquad
    \textbf{Majid Yazdani} \affilsup{1} \\%
    \textbf{Lambert Mathias} \affilsup{1} \qquad
    \textbf{Marzieh Saeidi} \affilsup{1} 
\\ 
    $^1$Meta AI \qquad
    $^2$Imperial College London \qquad
    $^3$University of Cambridge \\
    \texttt{nk2418@ic.ac.uk}\\
    \texttt{andreas.vlachos@cst.cam.ac.uk} \\
    \texttt{\{marzieh,mathiasl,myazdani\}@fb.com} 
}
\begin{document}
\maketitle
\begin{abstract}
Policy Compliance Detection (PCD) is a task we encounter when reasoning over texts, e.g. legal frameworks. Previous work to address PCD relies heavily on modeling the task as a special case of Recognizing Textual Entailment. Entailment is applicable to the problem of PCD, however viewing the policy as a single proposition, as opposed to multiple interlinked propositions, yields poor performance and lacks explainability.
To address this challenge, more recent proposals for PCD have argued for decomposing policies into expression trees consisting of questions connected with logic operators. Question answering is used to obtain answers to these questions with respect to a scenario. Finally the expression tree is evaluated in order to arrive at an overall solution.
However, this work assumes expression trees are provided by experts, thus limiting its applicability to new policies.
%
%
In this work, we learn how to infer expression trees automatically from policy texts. We ensure the validity of the inferred trees by introducing constrained decoding using a finite state automaton to ensure the generation of valid trees. We determine through automatic evaluation that 63\% of the expression trees generated by our constrained generation model are logically equivalent to gold trees. Human evaluation shows that 88\% of trees generated by our model are correct. 
\end{abstract}

\section{Introduction}

Reasoning over policies expressed in natural language is an important task in machine comprehension \cite{zhong-etal-2020-nlp}. The problem of determining the compliance of scenarios to written policies or other legal frameworks, referred to as policy compliance detection (PCD) \cite{saeidi-etal-2021-cross}  has a wide range of applications, from the legal domain, e.g. statutory law \cite{holzenberger-van-durme-2021-factoring} to the issue of content moderation on the social network websites \cite{Pavlopoulos-etal-2017-deeper}. An instantiation of PCD is the case in which a scenario is presented alongside a policy, 
and we need to determine
if a specific situation adheres to the rules defined in the text of the policy. 

Previous work \cite{holzenberger2020dataset} has formulated the PCD problem in the style of  Recognizing Textual Entailment \citep{dagan-etal-2005}, i.e. the answer is the inferred entailment relation between the policy (text) and the scenario (hypothesis). However, this RTE-based approach does not perform as well when applied to the PCD task because most policies have complex description, containing several connected propositions or clauses.

More recent work decomposes the policy into constituent propositions reformulated as questions, and uses a question answering (QA) model to obtain answers to each of these questions with respect to the scenario \cite{saeidi-etal-2021-cross}. Finally, an expression tree in propositional logic is employed to obtain a final answer. However, \citet{saeidi-etal-2021-cross} assumes  expression trees are provided by experts. This limitation makes the approach less straightforward to port to new domains and  policies.

\begin{figure}[ht]
    \centering
    \includegraphics[width=1\linewidth]{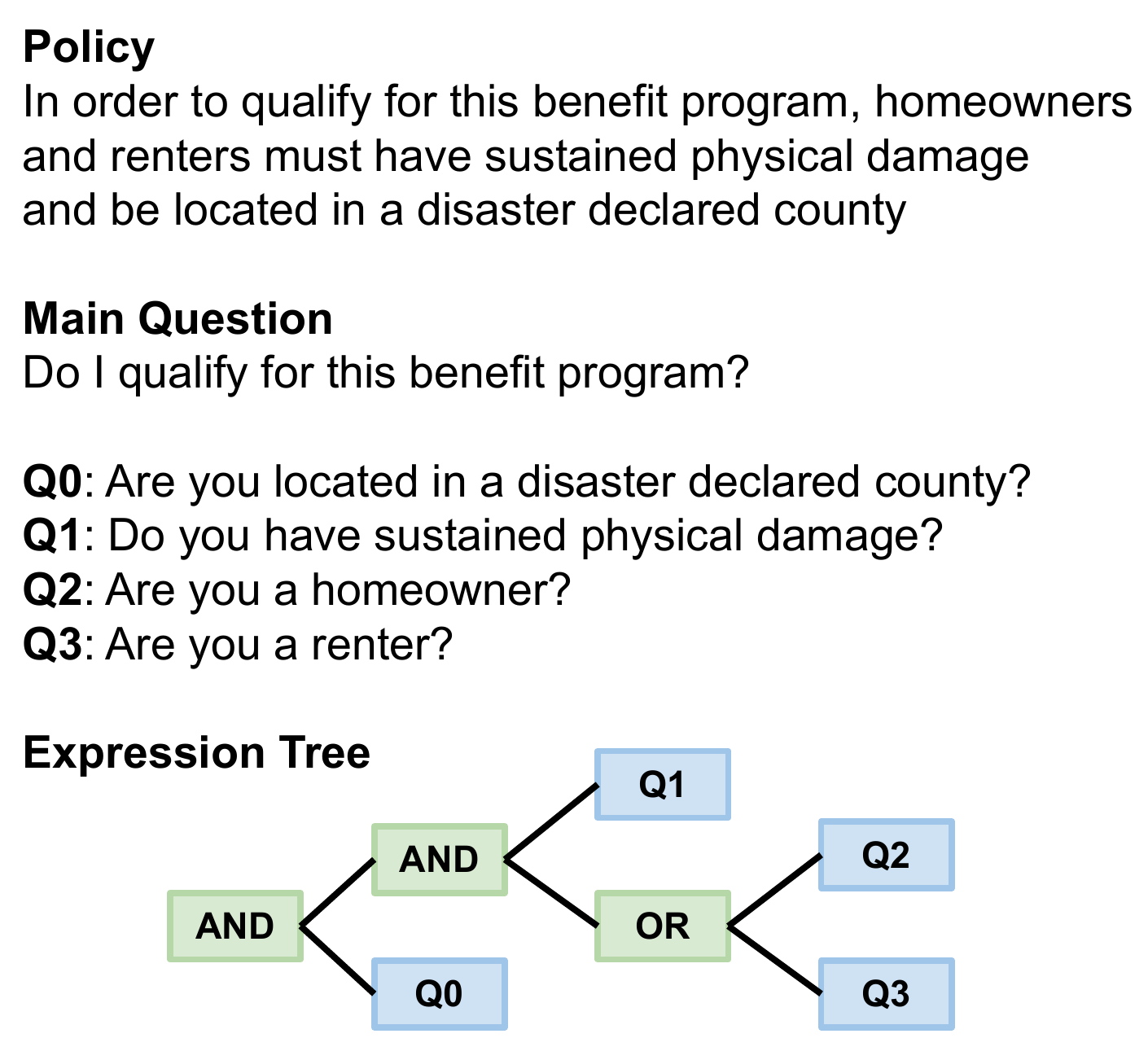}
    \caption{An expression tree which results from a policy being decomposed into four questions. The blue lines match the questions to the spans in the policy from which they are extracted. 
    }
    \label{fig:tree-inference}
\end{figure}

To address limitations associated with the cost of acquiring labelled trees for training, we seek to automate inferring the expression trees. To this end, we formulate this task as one of structured prediction \cite{kulmizev-etal-2019-deep}, i.e. we decompose policy texts into their constituent spans, rephrase these spans as questions, and use these questions to infer an expression tree. This results in a label prediction that is faithful to the answers to these questions, thus offering improved explainability. The expression tree summarises the constraints presented in the policy \cite{saeidi-etal-2021-cross} as shown in Figure \ref{fig:tree-inference}. 

We propose a two-stage approach to the problem: in the first stage, we generate all the necessary questions related to the policy and a given main question. In the second stage, given the policy, main question and the questions generated in the first stage, we output an expression tree. Our best-performing method models tree inference via auto-regressive sequence generation, using a finite state automaton to encode constraints relating to logic and generated questions, hence ensuring that generated trees are valid. 

We devise both automatic evaluation and human evaluation methods to assess the performance of our models. The auto-regressive sequence generation based model performs best, based on both automated and human evaluation metrics. It can also correctly generate more complex trees, i.e. those which decompose policies into multiple questions and contain multiple operators, than the neural scoring method can. The complexity of a tree is defined by the number of questions and number of unique operators it has in its expression.



\section{Expression Tree Generation}
\label{sec:methods}

\subsection{Task Definition}
Inferring an expression tree is the task of decomposing a policy into a set of summarising questions and generating a tree structure that connects those questions using logical operators \{\texttt{and}, \texttt{or}, \texttt{not}\}
to represent the reasoning process required to apply the policy text to a scenario.
Here, an input is a pair $\langle p, m\rangle$, where $p$ is a policy and $m$ is the main question, (see the example in Figure~\ref{fig:tree-inference}) and the output is an expression tree $t$. The expression tree can be expressed as a string containing question identifiers, e.g. \texttt{Q0}, logical operators, and parentheses. A tree should be a valid logic expression in propositional logic. For example ``\texttt{Q0 and not Q1}'' and ``\texttt{Q0 or (Q1)}'' are a valid expression trees, while ``\texttt{Q0 not Q1}'' and ``\texttt{Q0 or (Q1}'' are not valid expression trees.
Tree inference requires that we generate questions from the policy, and then infer the  relations between these questions, which we express in propositional logic, e.g. ``\texttt{not Q0 and Q1}".



There are  number of possible formulations which can be taken for tree inference.
We choose to perform this task in two stages: first we perform question generation, we then employ the generated set of questions as part of the input for the second stage of tree inference (see Figure \ref{fig:pipeline-text}). We choose this approach, rather than first inferring trees first and generating questions second, in order to exercise greater control over the number and content of questions generated.

\subsection{Question Generation}
Question generation itself is decomposed into the two steps described below.

\paragraph{Span Extraction} We first identify the constituent spans of the policy $p$ that are plausible question targets. We can have one or more spans in a policy, i.e. $ S = \{s_i, \dots, s_l\}$. We train a model to classify the \emph{question-worthiness} of each span, defined such that if a span is targeted by a question,  information  can be obtained that is necessary to answer the main question using the policy. 
For example, for the policy presented in Figure \ref{fig:tree-inference}, \textit{located in a disaster declared county} is a question-worthy span, which is rephrased as the question with identifier \texttt{Q3}, however the span \textit{In  order  to  qualify  for  this  benefit  program} is not question-worthy. 

We employ a split-and-rephrase \cite{narayan-etal-2017-split} text simplification approach in order train a text-to-text model to split policies into their constituent spans while preserving their meaning, so as to extract appropriate and well-formed spans. The spans generated through split-and-rephrase are filtered according to their question-worthiness, i.e.\ their relevance for forming questions to encapsulate the constraints represented in the policy. This is achieved by training a classifier for binary classification. This results in a set of spans $S'= \{s_i, \dots, s_{l^{'}}\}$, each span representing a self-contained sentence, such that $S' \subseteq S$ and $l' \leq l$.

\paragraph{Span Rephrasing} The second step is converting the spans into questions. The sentences which are evaluated as question-worthy are rephrased as questions. 
We generate a set of questions 
\begin{equation}
 Q = \{\hat{q}_{i=1}^{l'} \mid \arg\max_{q_{i}} P(q_i \mid p,m,s_i) \},
\end{equation}

such that each question in the set of generated questions $Q$ maximises the conditional likelihood given both the policy $p$, main question $m$ and the question-worthy span $s_i \in S$, as given by a trained auto-regressive model.


\subsection{Expression Tree Inference}
We then infer an expression tree $\hat{t}$ from $p$ and $m$, using $Q$. 
For tree generation, we represent the expression tree output as an infix logic expression e.g., \texttt{Q0 or Q1}. The tree summarises the constraints presented in the policy. We generate a tree, 
such that the generated tree $\hat{t}$ maximises the conditional likelihood given both the policy $p$, main question $m$ and the set of generated questions $Q$:
\begin{equation}
    \hat{t} =  \arg \max_{t} P(t \mid p,m,Q)
\end{equation}

\begin{figure}[t]
    \centering
    \includegraphics[width=0.9\linewidth]{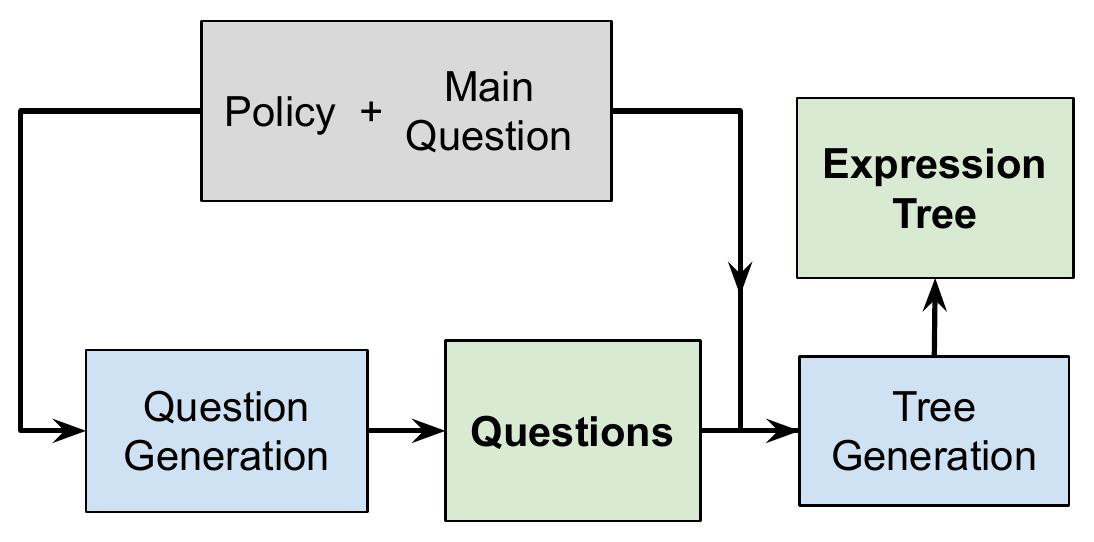}
    \caption{Our approach first generates questions based on the policy and main question. Policy and the main question and generated questions are then passed to the tree generation model.}
    \label{fig:pipeline-text}
\end{figure}

We explore two methods for tree inference: one which generates the expression tree auto-regressively, and a second method which is based on classification. 

\paragraph{Constrained generation for tree inference}

\begin{figure}[ht]
    \centering
    \scalebox{0.8}{
    \begin{tikzpicture}[>=stealth',shorten >=1pt,auto,node distance=2.6cm]
  \node[initial,state] (S0)      {$s_0$};
  \node[state] (S1) [right of=S0] {$s_1$};
  \node[state] (S2) [above of=S0] {$s_2$};
  \node[state] (S3) [above of=S1] {$s_3$};
  \node[state,accepting] (S4) [right of=S1] {$s_4$};
\path[->] (S0)  edge [bend left]    node {\texttt{[NOT]}} (S2)
                edge [bend right]   node [below] {\texttt{[Q]}}  (S1)
           (S2) edge  [bend right]             node [below] {\texttt{[Q]}}  (S1)
                edge  [bend left]     node {\texttt{[NOT]}} (S3)
           (S3) edge [bend left]               node {\texttt{[AND],[OR]}} (S1)
           (S1) edge [bend left]                node {\texttt{[Q]}} (S3)
                edge [bend right]              node [below] {\texttt{EOS}} (S4);
\end{tikzpicture}}
    \caption{State transitions for the finite state automaton encoded constrained tree generation. For brevity this diagram excludes the generation of parentheses.}
\label{fig:FSM}
\end{figure}
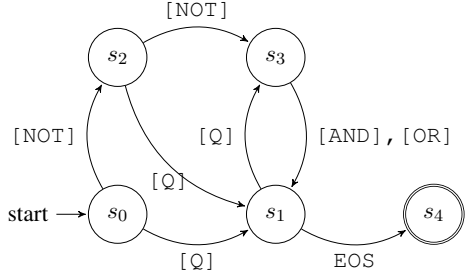

We use a finite state automaton to enforce constraints on the vocabulary when generating the tree in order to ensure the validity of the tree. The constraints employed for decoding are depicted in Figure~\ref{fig:FSM} and also outlined in detail in Algorithm~\ref{alg:cd-algorithm}. Note that we only output the question identifier tokens, e.g. \texttt{Q0}, for efficiency, as opposed to entire question texts.

We employ a greedy strategy for expression tree generation, whereby at each step we mask  (assign a large negative value to) 
the logits which correspond to tokens which would result in the generation of an invalid expression tree. To this end, we use a finite state automaton to encode the constraints for generating a valid expression tree (see Figure \ref{fig:FSM}). We use the current state of the finite state automaton to restrict the valid set of tokens, i.e.\ which logits should be masked at each step. The current state is influenced by the last generated, e.g.\ if we have 
generated ``\texttt{Q0 and}", then all but the logits for tokens \texttt{not}, \texttt{(}, and \texttt{Q1} are masked. 

Furthermore, the set of possible tokens which can be generated by the decoder are restricted to those corresponding to the logic operators \{\texttt{and}, \texttt{or}, \texttt{not}\}, parentheses tokens \{\texttt{(}, \texttt{)}\}, and the set of question identifier tokens $T_Q$, as determined by the number of generated questions, e.g., \{\texttt{Q0}, \texttt{Q1}, \texttt{Q2}\} for three generated questions. Once a question identifier token is generated, it is popped from the set $T_Q$ (Algorithm \ref{alg:cd-algorithm}, Line 12). As mentioned, we treat open parentheses in the same way as the other special tokens, i.e., the generation of open parentheses only differs from the generation of operators and question tokens so far as where they can be generated in the sequence according to the constraints defined by the finite state automaton. Tree generation terminates when all question tokens in $T_Q$ have been generated. We close any remaining open parentheses (Algorithm \ref{alg:cd-algorithm}, Lines 21-24), maintaining a stack to track opened parentheses during generation.

\begin{algorithm}[t]
\small
\begin{algorithmic}[1]
\REQUIRE generated questions $\mathbf{q} = q_0 \hdots q_n$

\ENSURE generated tree $\hat{t}$

\STATE {$\hat{t} \leftarrow \texttt{BOS}$} \hfill \COMMENT{$\blacktriangleright$ decoder output ids}

\STATE {$b \leftarrow 0$} \hfill \COMMENT{$\blacktriangleright$ track open parentheses}
\STATE $\texttt{$T_q$} \leftarrow \text{get all token ids for } \mathbf{q}$
\STATE $s \leftarrow \texttt{START}$\hfill \COMMENT{$\blacktriangleright$ track state, initialize with \texttt{START}}\\

\WHILE {$\texttt{$T_q$} \text{ is not empty} $}

\STATE{last\_token $\leftarrow \hat{t}$[-1])}\\
\STATE {$\mathbf{q} \leftarrow $\text{get new state given last token id}}
\STATE {$\text{valid\_tokens} \leftarrow \text{get\_valid\_tokens($\mathbf{q}$)} $}
\STATE {$\text{token} \leftarrow \text{generate new token with $\hat{t}$, valid\_tokens}$} 
\STATE {$\hat{t} \leftarrow \hat{t} \oplus \text{token} $}

\IF{$\text{token} \in {T_q} $}
\STATE {$T_q$.pop(token) \hfill \COMMENT{$\blacktriangleright$ pop question token from $T_q$}}
\ENDIF
\IF{$\text{token}$ \textbf{equals} $\texttt{[BR]} $}
\STATE {$b \leftarrow b + 1$}
\ENDIF
\IF{$\text{token \textbf{equals} } \texttt{[/BR]} $}
\STATE {$b \leftarrow b - 1$}
\ENDIF
\ENDWHILE
\WHILE {$b > 0 $}
\STATE {$\hat{t} \leftarrow \hat{t}$ $ \oplus $ \texttt{[/BR]} }
\STATE{$b \leftarrow b - 1$}
\ENDWHILE
\RETURN decode($\hat{t}$ $\oplus$ \texttt{EOS})
\end{algorithmic}
\caption{Constrained expression tree decoding.}
\label{alg:cd-algorithm}
\end{algorithm}

\paragraph{Neural scoring for tree inference} 
In the first method, we devise a neural scoring approach, inspired by the structured perceptron \cite{collins-2002-discriminative,huang-zhao-2015-scalable}. We train a model for pairwise compatibility between a policy and tree and use this model to predict the most likely tree given the number of generated questions and the policy text. 
In this classification approach to tree inference, for training the model we construct a dataset for training where the positive examples are gold instances from the training data. In order to generate negative training examples, we match policies and gold questions to trees which are inaccurate, but represent the same number of questions. At test time, the model takes as input the policy text, main question, and the set of generated questions. For each possible expression tree, which we consider to be all the expression trees in the training data whose number of questions is equal to the number of generated questions. For each possible tree we output a compatibility score, i.e. a measure of how well the expression tree represents the policy. We rank the expression trees, and chose that which has the highest compatibility score.

\section{Dataset}

Our dataset is an extension of the QA4PC dataset for cross-policy compliance detection \cite{saeidi-etal-2021-cross}. We describe our dataset annotation process, and we also present our analysis, including key statistics, related to the dataset. 

\paragraph{Dataset Annotation}
QA4PC provides annotations of expression trees for its test and development sets. To annotate expression trees for the training set, we started with the annotation guidelines from QA4PC. Two expert annotators (UK nationals) initially annotated 100 examples. They reviewed each other's annotations and discussed the disagreements. The remaining instances were then annotated by a single expert after discussion. 

\paragraph{Dataset Statistics}
\label{sec:data-analysis}
QA4PC-trees, together with the QA4PC, consists 
of 447 expression trees for training (annotated by us) and 193 for development and test (from QA4PC). 42.7\% of the expression trees in the training data decompose their respective policies into only a single question and zero (i.e.\ ``\texttt{Q0}'') or one (i.e. ``\texttt{not Q0}'') operator. 8.9\% of expression trees in the training data consist of both \texttt{and} and \texttt{or} operators, e.g. ``\texttt{Q0 and Q1 and not (Q2 or Q3)}.''



\section{Experiments}



\subsection{Question Generation}

For question generation, we consider two baselines: a pattern-based method and one which employs regular expressions. 

\paragraph{Pattern-based} Regular expressions are used to identify declarative spans in policies. These regular expressions extract bullet-pointed spans and spans which follow wording such as ``must be'' and ``should apply''. Note that we only use the policy text and not the main question for this approach. For question rephrasing, we consider a number of patterns for transforming declarative phrases into questions, e.g. \textit{it was} would become \textit{was it}) considering twenty-six phrases in total. If the sentence in question does not match one of these predefined transformations, we prefix the span \textit{Do you have} to the question if it is the start of a sentence, and we append ``Are you'' to the start if it does not. The choice as to which prefix is appended is determined by examination of the most probable prefix given the training instances.

\paragraph{Neural span extraction and rephrasing} For the second baseline, we employ a method for parsing the policy text to extract spans, whereby we extract bullet-points as spans, and use pattern-matching to extract as spans sentences which contain specific phrases, e.g. \textit{must}, \textit{would}. Each selected span in the training data is then evaluated for question-worthiness according to pair-wise with respect to the gold question, using Sentence-BERT and a threshold which is set experimentally using a development set (10\% of the training data). This is used to fine-tune a BART model for question-worthy classification. In addition, we fine-tune a BART model for question rephrasing.

\subsection{Split-and-Rephrase \& neural rephrasing}

For this, we use T5 model pre-trained \cite{DBLP:journals/jmlr/RaffelSRLNMZLL20}, on the WikiSplit dataset  \cite{botha-etal-2018-learning}. 
We infer the suitability of candidate sentences in the following way. 
First, we use automatic means to annotate the constructed sentences in the training set for suitability - assigning each sentence a binary score based on its semantic similarity to each of the gold questions. We use the pre-trained Sentence-BERT model \cite{reimers-gurevych-2019-sentence} for this purpose. We choose the threshold of 0.65 for determining suitability of the sentences and we choose this threshold experimentally, by evaluating the performance of threshold values on a randomly sampled 10\% split of the training data. In addition, for policies which include bullet points, we make the assumption 
that all bullet points represent relevant questions. Hence all sentences formed from bullet points are also annotated as suitable. 
Using these annotations, we fine-tune a classifier that determines the suitability of a sentence. The classifier is a BART language model~\cite{lewis-etal-2020-bart}. 

The classifier takes as input the policy, main question and the candidate sentence, as generated by the split-and-rephrase method. We then train a generator model, i.e. BART, in order to rephrase each suitable sentence into a well-formed question. We train the question generator using the policy, the main question and the candidate sentence as input, and the gold question as the output. At test time, for instances for which no suitable sentences are predicted, we use the sentence with the highest suitability score to form a singleton question set.

\subsection{Tree Inference}
We employ two baselines for tree inference: a pattern-based method and a random baseline method.

\paragraph{Pattern-based} For pattern-based tree inference, a series of hand-crafted rules, formed by inspecting of the structure of policies in the training data, are applied. For pattern-based tree inference, the generated questions are ignored, it is based solely on the text of the policy. For policies which consist of bullet-pointed statements, the questions associated with these bullet points are assumed to be a conjunction of literals, e.g. ``\texttt{Q0 and Q1}" if the sentence immediately preceding the bullet points contains \emph{must} or \emph{both}. Otherwise, the bullet pointed questions are interpreted as a disjunction of literals,  i.e. ``$\texttt{Q0 or Q1}$". Furthermore, if a negation (\emph{not} or '\emph{nt}) appears in the sentence preceding the bullet points, we enclose the logical expression associated with the bullet pointed text in a negation, e.g. ``\texttt{Q0 and Q1}'' would become ``\texttt{not (Q0 and Q1)}".  In addition to this, all other logical operators between literals in the expression tree are assumed to be the \texttt{and} operator.

\paragraph{Random and most common trees}  For a given question set, we select at random a tree which contains a number of questions equal to the size of the set of generated question. For illustration, if question generation yields the following question set  $\{\texttt{Q0}: \textit{``Are you eligible?''}\}$, then a tree will be chosen at random from the following possible trees $\{``\texttt{not Q0}", ``\texttt{Q0}"\}$. Furthermore, we introduce an additional baseline for which, given a number of questions, we choose the tree from the training data representing the most frequently occurring tree which contains the same number of questions.

\subsubsection{Constrained tree generation}
For this method, we employ a T5 model for tree generation. Our model's inputs are the set of questions generated in the previous stage, in addition to the policy and the main question (see Figure \ref{fig:pipeline-text}). 
During training, we encode operators as (\texttt{[OR], [AND], [NOT]}), question identifiers (\texttt{[Q0]} - \texttt{[Q9]}) and parentheses as additional special tokens (\texttt{[BR], /BR]}). 
For example, given a policy and main question $p, m$, and generated question sequences $q_0$ and $q_1$, we encode the input sequence as follows: ``
    \underline{input}: $p$ \underline{context}: $m$ \texttt{[Q0]} $q_0$ \texttt{[Q1]} $q_1$ '', a possible output is ``\texttt{[Q0] [AND] [Q1]}''.
    
\subsubsection{Neural scoring tree inference}

For training, we create use each policy and its annotated expression tree as a positive example. For the negative examples, we match incorrect trees to policies with the same number of questions. We sample an equal number of positive and negative examples for training the compatibility model. For this method we fine-tune a single BART model. 

As test time, for a policy, we predict its compatibility with each candidate tree in the training data that contains the same number of questions, as the size of the set of generated question set for the policy. We also reduce the number of possible trees by treating logically equivalent trees as a single class e.g., in the case where we are considering possible trees which contain three questions we would treat \texttt{Q0 and Q1 and Q2}, \texttt{Q1 and Q2 and Q0}, and all other equivalent expressions as a single case. Once we have evaluated the compatibility of all policy-tree pairs, we rank the trees and select the highest ranked expression tree as the output for this method. Furthermore, at test time, we skip question sets which include a number of questions for which we have no examples in the training data, e.g. no trees summarise more than eight question.



\subsection{Data Augmentation}

In order to compensate for the small number of training examples, especially those with complex expression trees, we employ a variety of data augmentation methods. 
We implement four strategies for augmentation: modifying questions in policies, this amounts to splitting questions and reflecting these changes in the expression tree; substituting expression trees for logically equivalent but non-identical trees; substituting conditional phrases in policies, e.g. \textit{for all}, and reflecting this in the tree; and finally omitting bullet points in policies and reflecting this in the question set and tree.
Through our augmentation methods, we increase the number of training samples by 645, which takes the number of training examples up an order of magnitude to a total of 1,092.
We employ this augmentation approach for all trainable tree inference methods.

\begin{table*}[ht]
    \centering
    \begin{tabular}{lccccc}
    \hline
        \textbf{Model} &  \textbf{BLEU-1} & 
        \textbf{BLEU-4} & \textbf{S-BERT} & \textbf{ROUGE$_{\text{L}}$} & \textbf{METEOR}\\
    \hline
     \text{Pattern-based SE \& rephrasing} & 49.48 & 43.80 & 70.51 & 50.98 & 60.20 \\

     \text{Neural SE \& rephrasing} 
    & 67.01 & 53.73 & 72.27 & 60.13 & 53.86\\
    \hline

    \text{Split-\&-Rephrase \& neural rephrasing} & \textbf{76.76} & \textbf{68.39} & \textbf{83.55} & \textbf{72.40} & \textbf{69.82} \\
    \hline
   \text{Annotated spans as questions} & 73.33  & 68.42 & 84.02 & 74.58 & 51.42\\
    \hline
    \end{tabular}
    \caption{Automatic evaluation of question generation methods (span extraction (SE) and rephrasing),
    using BLEU-1, BLEU-4, ROUGE$_{\text{L}}$, METEOR metrics, and S-BERT pair-wise similarity between generated and gold questions. 
    We present annotated spans as an upper bound for question generation, for which we sample 60 test data instances.}
    \label{tab:question-evaluation}
\end{table*}

\section{Results}
\subsection{Automatic Evaluation}

\paragraph{Questions} We use automatic evaluation metrics to assess the performance of both the question generation and also the tree generation models. We employ BLEU  \cite{papineni-etal-2002-bleu}, ROUGE$_{\text{L}}$ \cite{lin-2004-rouge}, and METEOR \cite{banerjee-lavie-2005-meteor} for automatic question evaluation. Furthermore, we use Sentence-BERT, which we also employ for question generation (see Section \ref{sec:methods}), to compute the semantic similarity between the generated and gold questions. For each generated question, we compute the maximum score between the question and each of the gold questions. Results for question generation evaluation are shown in Table \ref{tab:question-evaluation}. We include spans which we manually annotated for the purpose of evaluation as an upper bound for question generation. As we can see, split-and-rephrase shows considerable improvement over the use of pattern matching for splitting policies.

\paragraph{Trees} We also use two automatic metrics for evaluating the accuracy of the generated trees. The first is the proportion of the generated trees that are \emph{identical} to the gold tree associated with the same policy. The second metric looks to evaluate where the truth tables for the two expression trees are \emph{equivalent}, assuming the number of questions present in both the generated and gold expressions are the same; e.g. the two expression trees ``\texttt{not Q0 and not Q1}'' and ``\texttt{not (Q0 or Q1)}'' are equivalent, but not syntactically identical (see Table \ref{tab:tree-evaluation}).

\begin{table}[ht]
    \centering
    \begin{tabular}{lccc} 
    \hline
    &  \textbf{Identical} & \textbf{Equivalent}\\
    \hline
    \text{Random}  &   0.207 & 0.260 \\
    \text{Most common}  & 0.458 & 0.474\\
    \text{Pattern-based}    & 0.285 & 0.363\\
    \hline
   \text{Generation}    & \textbf{0.513} & \textbf{0.627}  \\
   \text{Neural Scoring} & 0.319 & 0.387 \\
   \hline
    \end{tabular}
    \caption{Automatic evaluation for expression tree inference methods with gold questions. }
    \label{tab:tree-evaluation}
\end{table} 

\subsection{Policy Compliance Detection Evaluation}

We conduct end-to-end evaluation (tree generation and then policy compliance detection (PCD)) for both the constrained tree generation and the neural scoring methods. The results for end-to-end evaluation on the  downstream task of the PCD task are presented in Table~\ref{tab:end-to-end-results}.

For PCD evaluation, we first align generated questions to gold questions using the a S-BERT derived semantic similarity score between each generated question and all the gold question in a policy.  This is because our question generation can produce questions in different order, e.g. for the logic expression  ``\texttt{Q0 and not Q1}'', the generated question corresponding to the gold ``\texttt{Q0}'' might be associated with the question identifier ``\texttt{Q1}''. We then run a QA model, following the same approach taken in \cite{saeidi-etal-2021-cross},  on for each generated question, given a scenario in the test set of QA4PC. We then combine the answers to all the generated questions from the scenario, based on the predicted expression tree of the policy to get the final answer.

\begin{table}[ht]
    \centering
    \begin{tabular}{lcc}
        \hline
        \textbf{Method} &  \textbf{Macro acc.} & \textbf{Micro acc.}\\
        \hline
        \text{Entailment} & 0.44 & --\\
        \text{Gold trees} & 0.70 & -- \\ 
        \text{Pattern-based} & 0.47 &  0.48\\
        \hline
        \text{Generation} & \textbf{0.47} & \textbf{0.50}\\
        \text{Neural Scoring} & 0.45 & 0.46\\
        \hline
    \end{tabular}
    \caption{PCD evaluation of inferred trees. The entailment and pattern-based methods are baselines. Further gold trees serve as an upper bound. 
    }
    \label{tab:end-to-end-results}
\end{table}

\subsection{Human Evaluation}
To better assess the quality of the generated questions and expression trees, we manually evaluate the predictions for all the instances in the test set. We do this for the two best-performing approaches -- neural scoring and constrained generation. 
\paragraph{Procedure} We ask each evaluator to judge each generated question on its fluency and intelligibility. A question is fluent if it is grammatically correct. A question is intelligible if the meaning is clear, even though it may not be grammatically correct, e.g. \textit{Is you over 25?}.  We also ask evaluators to identify which questions, if any, in the gold question set is aligned to the generated question. Note that a generated question can be a combination of gold questions, e.g. a generated question can be aligned to ``\texttt{Q0 or Q1}'' in the gold set. We consider two measures of alignment: exact and fuzzy.  Exact alignment is when a generated question is semantically equivalent to a gold question (for the same policy). In contrast the exact alignment, fuzzy alignment is where the generated question can be partially aligned to a gold question but some of the information is missing. We want to be able to reward the model's effort, even if the generation is not perfect. Further, given the passage, main question and the generated questions the human evaluators, need to decide whether the expression tree is correct.
\paragraph{Quality of Human Evaluation} We ask two human evaluators (authors of the paper, not the first author who was involved in the design and implementation of the experiments) to evaluate 50 test instances. We calculate the inter-annotator agreement. Since the inter-annotated agreement is quite high, a single annotator then perform evaluation for all the remaining instances for both approaches. The agreement scores can be found in Table~\ref{table_agreement}. We calculate the percentage of the time that human evaluators agreed on a decision, i.e.\ whether a question is fluent. Since a majority of the generated questions are fluent and intelligible, there is a high imbalance between labels that evaluators are assigning (mainly 1) and therefore, Cohen's $\kappa$\footnote{\url{https://en.wikipedia.org/wiki/Cohen\%27s_kappa}} \cite{cohen1960coefficient} presents limitations.

\begin{table}[ht]
\centering
\scalebox{0.95}{
\begin{tabular}{l  c}
\hline
\textbf{Measures} &  \textbf{Agreement \%} \\
\hline
 \text{Fluency} & 88 \\
 \text{Intelligibility}& 93 \\  
\text{Exact Alignment} & 77 \\
 \text{Fuzzy Alignment} & 89 \\
 \text{Tree} & 94\\
 \hline
\end{tabular}
}
\caption{
Agreement of human evaluators.} \label{table_agreement}
\end{table}

\paragraph{Human Evaluation Results}

\begin{table}[ht]
\centering
\begin{tabular}{llll}
\hline
     \textbf{Fluency} &
     \textbf{Intelligibility} &
     \makecell{\textbf{Exact}\\\textbf{Align.}} &
     \makecell{\textbf{Fuzzy}\\\textbf{Align.}} \\
     \hline
     92\% & 94\% & 79\% & 88\%\\
     \hline
\end{tabular}
\caption{Human evaluation of fluency and intelligibility of generated questions. We also identify whether a generated question is aligned with a gold question(s).
}
\label{tab:human-evaluation-questions}
\end{table}

\begin{table}[ht]
\centering
\begin{tabular}{lll}
\hline
     \textbf{Method} &  \textbf{Tree} 
     \\
\hline
     \text{Generation} &  88\% (85/96) 
     \\
     \text{Neural Scoring} & 54\% (52/96) 
     \\
\hline

\end{tabular}
\caption{Frequency with which valid trees are inferred the from generated question (out of 96 aligned trees). Also PCD micro and macro accuracy for correctly inferred trees, for same subset of 53 instances.
}
\label{tab:human-evaluation-trees}
\end{table}



Table~\ref{tab:human-evaluation-questions} shows the percentage of the times that the human evaluator labelled generated questions as fluent or intelligible. The generated question may lack some information which is present in the gold question. Table~\ref{tab:human-evaluation-trees} shows the percentage of times that the human evaluator labelled the generated tree as correct. Note that it only makes sense to judge the correctness of a tree if all of its questions is aligned to gold questions and vice versa. We call such trees \emph{aligned} trees.

\begin{figure}[t]
    \centering
     \begin{subfigure}[b]{0.42\textwidth}
         \centering
         \includegraphics[width=\textwidth]{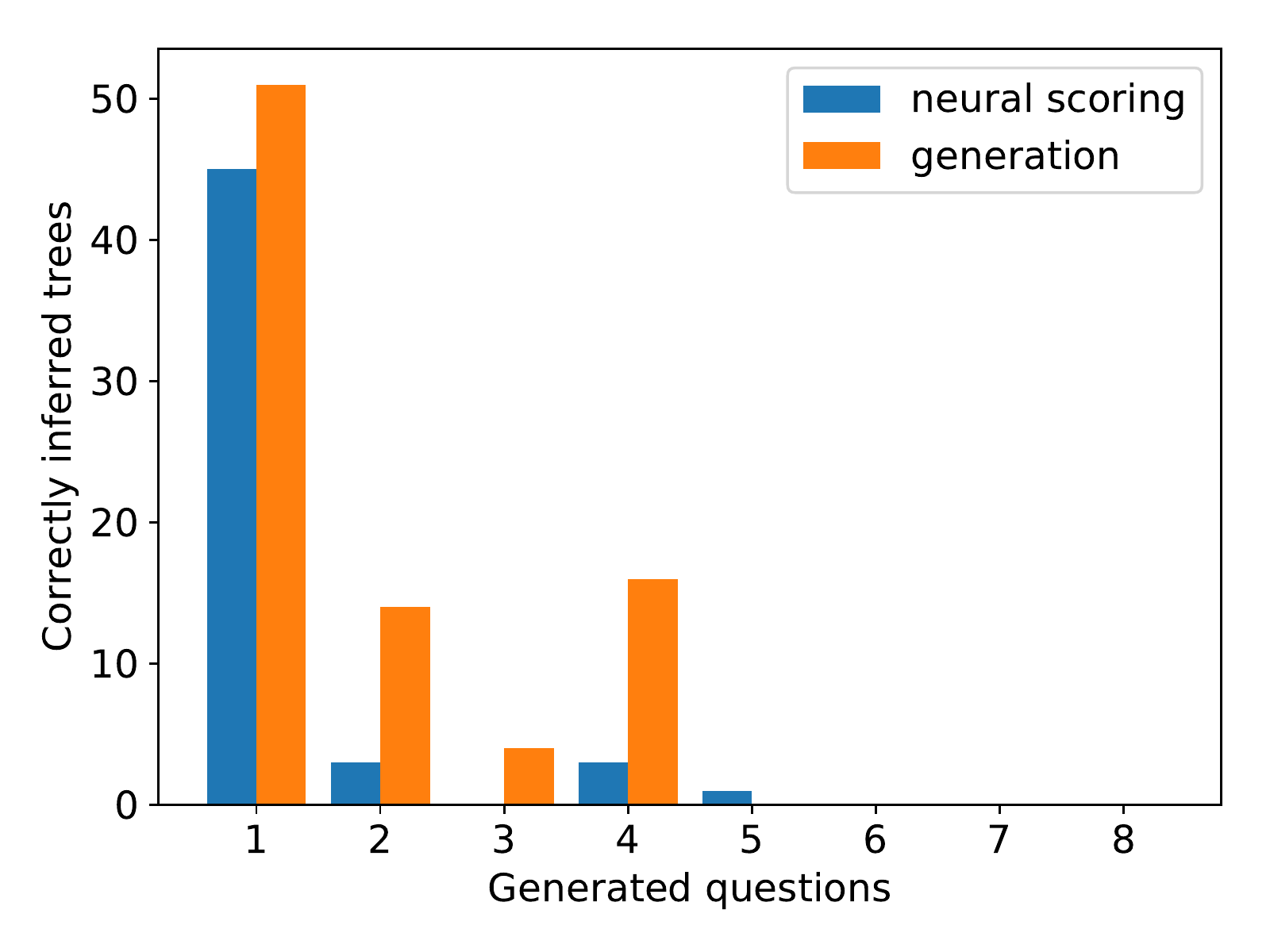}
         \caption{Complexity by number of questions generated.}
         \label{fig:complexity-questions}
     \end{subfigure}
     \begin{subfigure}[b]{0.42\textwidth}
         \centering
         \includegraphics[width=\textwidth]{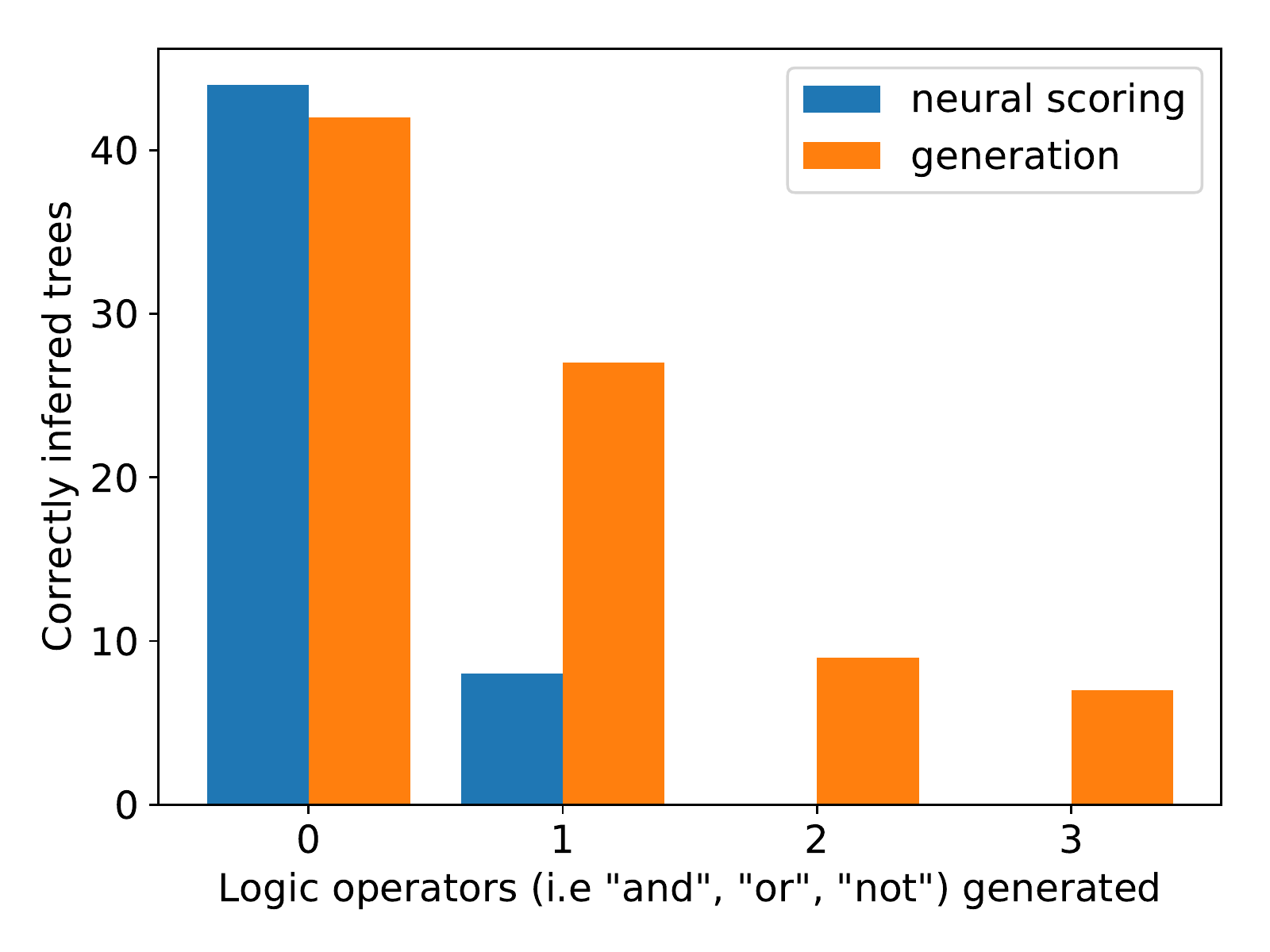}
         \caption{Complexity by the number of operators generated.}
         \label{fig:complexity-operators}
     \end{subfigure}
    \caption{Histogram of the complexity of trees by number of questions (top) and number of unique operators (bottom) in the inferred expression tree.}
    \label{fig:complexity-histograms}
\end{figure}

Figures~\ref{fig:complexity-histograms} shows the complexity of the correctly generated trees using the generation and the neural scoring method, by the number of generated questions and by the number of \textit{unique} operators (including parentheses) in the expression tree. As Figure \ref{fig:complexity-histograms} shows, the generation method can generate correct trees of higher complexities. Whereas the neural scoring method mainly generates single question trees, i.e. ``\texttt{Q0}'' and ``\texttt{not Q0}'' or trees which only include one type of operator, e.g. ``\texttt{Q0 or Q1 or Q2}''. The generation method infers trees such as ``\texttt{not (Q0 or Q1)}.''

\section{Related Work}



\paragraph{Question Generation} An overwhelming majority of the literature on question generation either concerns facts, i.e. the purpose of question generation in these cases is to extract fact-based sentences from a text, and identify the answer and question portions of these sentences. Question generation \cite{heilman-smith-2010-good} of this form is typically employed as part of a pipeline for question-answering \cite{duan-etal-2017-question,wang-etal-2020-pathqg,lewis-etal-2021-question} or for generating questions for fact-checking summary briefs  \cite{hassan2017claimbuster,jimenez2017towards,fan-etal-2020-generating}.  

We present a special case of generation, i.e. as a preliminary step in structured prediction, i.e. tree inference. Our task formulation differs from other approaches as we perform question generation which is entirely extractive in nature, i.e. we seek only to extract questions from the policy text, which are prerequisites for the main question. In some cases, generation requires the extraction of answers to the questions, as well as the questions themselves from an input passage, i.e. \textit{question generation for QA}. For tree inference, we are only tasked with extracting questions. However, our case may be considered a special case of question generation for QA, where the \emph{main question} itself is the answer to our generated questions. Furthermore, in our case, policies are decomposed into questions to provide a more explainable method for PCD as compared to entailment, task-specific explainablility has been explored in NLP \cite{kotonya-toni-2020-explainable,wiegreffe-marasovic-2021-review}.

\paragraph{Structured Prediction \& Tree Inference}

Structured prediction describes a number of techniques for predicting structured outputs, e.g. trees from input data \cite{smith2011linguistic}.
Syntactic parsing is an application of structured prediction in NLP, for which a number of approaches have been formulated \citep{dong-lapata-2016-language,10.1162/COLI_a_00158}. Although, some techniques employed in these works are not necessarily amenable to our problem of generating expression trees, there is still some overlap, one of them being the use of generative models for this task.
\section{Conclusion and Future Work}

We employ a novel finite state automaton constrained generation approach to ensure valid  expression trees, which outperforms the entailment approach. There are a number of potential avenues for further research, one would be to construct more expressive trees for policy summarisation, e.g. by using predicate logic as opposed to propositional logic. It would be interesting to explore strategies for tree inference with the goal of maximising performance on the downstream PCD task.




\bibliography{anthology,custom}
\bibliographystyle{acl_natbib}

\end{document}